%% file: main.tex
\pdfoutput=1

\documentclass[11pt]{article}

\newcommand\blfootnote[1]{%
  \begingroup
  \renewcommand\thefootnote{}\footnote{#1}%
  \addtocounter{footnote}{-1}%
  \endgroup
}

\usepackage[final]{acl}

\usepackage{times}
\usepackage{latexsym}
\usepackage{comment}
\usepackage{color}
\usepackage{listings}
\usepackage[most]{tcolorbox}
\usepackage{lipsum} 

\usepackage{colortbl}
\usepackage[T1]{fontenc}

\usepackage[utf8]{inputenc}

\usepackage{microtype}
\usepackage[most]{tcolorbox}
\usepackage{inconsolata}

\usepackage{mathtools}
\usepackage{makecell}

\usepackage{array,multirow,graphicx}
\usepackage{amsmath}
\usepackage{amssymb}
\usepackage{booktabs}
\usepackage{algpseudocode}

\usepackage[export]{adjustbox}
\usepackage{float}

\usepackage[capitalize]{cleveref}
\crefname{section}{Sec.}{Secs.}
\Crefname{table}{Table}{Tables}

\tcbset{
    mypromptbox/.style={
        enhanced,
        sharp corners,
        boxrule=0pt,             
        colframe=white,          
        colback=gray!10,         
        coltitle=black,
        fonttitle=\bfseries,
        attach boxed title to top left={
          xshift=0mm, yshift=-2mm
        },
        boxed title style={
          colframe=white,
          colback=white,
          boxrule=0pt,
          sharp corners,
          left=0mm, right=0mm, top=0mm, bottom=0mm,
        },
        before skip=10pt,
        after skip=10pt,
        left=2mm, right=2mm, top=2mm, bottom=2mm,
    }
}
  
\input{commands/mathcommand}

\input{commands/mycommand}

\newcommand{\myalg}{\texttt{\textbf{CIFLEX}}\xspace}

\newcommand{\mytitle}{\myalg: Contextual Instruction Flow for Sub-task Execution in Multi-Turn Interactions with a Single On-Device LLM}

\title{\mytitle}

\author{Juntae Lee$^\dag$\hspace{1em}Jihwan Bang$^\dag$\hspace{1em}Seunghan Yang$^\dag$\hspace{1em}Simyung Chang$^\ddag$\\
{Qualcomm AI Research$^*$$^\dag$} \\ 
{\texttt {\small\{juntlee, jihwbang, seunghan\}@qti.qualcomm.com}}}

\begin{document}
\maketitle

\blfootnote{\hspace{-1.8em}$^*$Qualcomm AI Research is an initiative of Qualcomm Technologies, Inc.}
\blfootnote{\hspace{-1.8em}$^\ddag$Work completed while employed at Qualcomm AI Research.}
 
\begin{abstract}

We present \myalg (\underline{\textbf{C}}ontextual \underline{\textbf{I}}nstruction \underline{\textbf{FL}}ow with \underline{\textbf{EX}}ecution), a novel execution system for efficient sub-task handling in multi-turn interactions with a single on-device large language model (LLM). As LLMs become increasingly capable, a single model is expected to handle diverse sub-tasks that more effectively and comprehensively support answering user requests.
Naive approach reprocesses the entire conversation context when switching between main and sub-tasks (e.g., query rewriting, summarization), incurring significant computational overhead. \myalg mitigates this overhead by reusing the key-value (KV) cache from the main task and injecting only task-specific instructions into isolated side paths. After sub-task execution, the model rolls back to the main path via cached context, thereby avoiding redundant prefill computation. To support sub-task selection, we also develop a hierarchical classification strategy tailored for small-scale models, decomposing multi-choice decisions into binary ones. Experiments show that \myalg significantly reduces computational costs without degrading task performance, enabling scalable and efficient multi-task dialogue on-device.

\end{abstract}
\input{Intro}

\input{Method}

\input{Exp}

\input{Related}

\section{Conclusions}
\vspace{-0.2cm}
We introduced \myalg, a new multi-turn interactions strategy that enables efficient multiple sub-task handling under a single on-device LLM. \myalg reduces redundant computation by reusing KV cache from the main conversation flow and executing sub-tasks through instruction-only prefilling in side paths. 
After execution, the model seamlessly returns to the main path using the preserved cache, eliminating costly prefill operations. To support sub-task selection, we proposed a hierarchical classification strategy that decomposes multi-choice decisions into binary ones, making it suitable for small-scale models. For validation, we present novel multi-turn, multi-task datasets (TopiOCQA-Task+ \& QReCC-Task+). Experiments demonstrated that \myalg notably lowers computational cost while maintaining task performance, highlighting its practicality for scalable, on-device conversation systems.
\vspace{-0.25cm}

\section{Limitations} 
\label{sec:limitation}
While \myalg demonstrates strong efficiency gains and robust task handling within multi-turn conversations, several limitations remain in multi-turn conversation system. Due to the current limitations of LVMs, results from image or audio API calls are approximated using textual descriptions in LLM. For retrieval sub-tasks, we rely on off-the-shelf retrievers. Future advances in the related fields may enable seamless integration as a complete multi-turn conversation with AI assistants. Also, our datasets, while covering multi-turn interactions, could be extended to include further longer conversational trajectories. Lastly, although \myalg significantly reduces prefill latency, generation latency remains a dominant cost, which could be mitigated through further studying on the optimization in future edge hardware like LRU caching.



\bibliography{bib}

\newpage
\input{appendix}

\end{document}

%% file: commands/mathcommand.tex
\usepackage{amsmath,amsfonts,bm}

\def\eqref#1{equation~\ref{#1}}

\def\1{\bm{1}}

\DeclareMathAlphabet{\mathsfit}{\encodingdefault}{\sfdefault}{m}{sl}
\SetMathAlphabet{\mathsfit}{bold}{\encodingdefault}{\sfdefault}{bx}{n}



%% file: commands/mycommand.tex
\usepackage{geometry}
\usepackage{booktabs} 
\usepackage{bbm}
\usepackage{mathtools}
\usepackage{nccmath}
\usepackage{setspace}

\usepackage{caption}
\usepackage{subcaption}

\usepackage[linesnumbered,ruled,vlined]{algorithm2e}

\SetKwInput{KwInput}{Input}                
\SetKwInput{KwOutput}{Output}              

\SetCommentSty{mycommfont}

\SetAlCapSty{algcapsty}

\usepackage[T1]{fontenc}
\usepackage{wrapfig,lipsum,booktabs}

\usepackage{soul}
\usepackage{dsfont}
\usepackage{enumerate}
\usepackage{enumitem}

\usepackage{amsmath}
\usepackage{amsfonts}
\usepackage{bbm}
\usepackage{dsfont}
\usepackage[Symbol]{upgreek}
\usepackage{lscape}
\usepackage{caption}
\usepackage{balance}
\usepackage{xspace}
\usepackage{float}
\usepackage{kotex}

\usepackage{wasysym}
\usepackage{xcolor}
\usepackage{multirow}
\usepackage{array, boldline, rotating}

\usepackage{amssymb}
\usepackage{pifont}
\renewcommand*\eqref[1]{(\ref{#1})}

\newcommand{\yes}[1]{\textcolor{blue}{[YES]}}
\newcommand{\no}[1]{\textcolor{orange}{[NO]}}
\newcommand{\na}[1]{\textcolor{gray}{[N/A]}}
\newcommand{\eg}{\emph{e.g.,~}}

\definecolor{LightCyan}{rgb}{0.88,1,1}
\definecolor{Blue}{rgb}{0, 0.5, 1}
\definecolor{Green}{rgb}{0.0, 0.8, 0.0 }
\definecolor{Red}{rgb}{0.95, 0.55, 0.6}
\definecolor{Skyblue}{rgb}{0.6, 0.6, 0.95 }

\NewDocumentEnvironment{suptitle}{ +b }{
    \twocolumn[{#1}]%
}{}

\NewDocumentCommand{\supptitle}{s}{
\begin{suptitle}
        \centering
        \rule{\textwidth}{0.03cm}\\[0.1cm]
        {\Large 
            \textbf{- Appendix -} \\ [0.2cm]
        }
        {\Large 
            \textbf{\mytitle }
        }\\
        \rule{\textwidth}{0.03cm}\\[0.2cm]
\end{suptitle}}

%% file: Intro.tex
\section{Introduction}
As Large language models (LLMs)~\cite{dubey2024llama,achiam2023gpt,yang2024qwen2,jiang2023mistral} engage in more multi-turn interactions~\cite{hassan2024advancing,guan2025evaluating,laban2025llmslostmultiturnconversation}, user poses various and complex requests. As such, a variety of sub-tasks, such as query rewriting~\cite{zhang-etal-2024-adaptive,Zxinbei_query}, API call~\cite{yao2023reactsynergizingreasoningacting} or complex reasoning~\cite{singh2025agenticreasoningtoolintegration}, are necessary for better supporting user request in the middle of a multi-turn interaction session. Moreover, the LLM should keep aware of the historical context of the multi-turn conversation for correct sub-task executions as well as reliable answer for the main request. However, in edge devices, while maintaining the context of conversation in a main model, deploying additional side-task specific models for every sub-task is impractical due to its memory and compute constraints. Encouragingly, LLMs progress at a remarkable pace, and thus a single LLM is expected to handle a variety of tasks.

Despite the rapid progress of the versatility of a single LLM, it has paid limited attention to the efficiency of the scenarios where sub-task execution and decision-making operate \textit{over long, multi-turn} conversations under an LLM. Explicitly including sub-task execution within the conversation history leads to not only excessively long contexts but also exposes unnecessary internal processing steps to the user. Then, as a naive approach, one could perform sub-task execution by reloading the entire conversation history and inserting new task-specific instructions whenever switching from the main task to a sub-task. However, this incurs substantial computational and memory overhead due to repeated prefill operations, which is particularly problematic for on-device deployment. For example, in a conversation of around 20 turns, the prefill required can be 300 times more tokens than the generation. As the number of turns grows, the cost of managing context and instructions across multiple tasks becomes a dominant bottleneck, especially when such operations are repeated throughout a session.

Decoder-based LLMs~\cite{dubey2024llama,achiam2023gpt,yang2024qwen2,jiang2023mistral} can alleviate redundancy during generation via key-value (KV) caching, which stores intermediate token representations to avoid recomputation over previously seen inputs. Recently, several works~\cite{lyu2025streamliningcollaborativechainmodels,liu2024droidspeakkvcachesharing} have attempted to exploit KV cache sharing across differently fine-tuned models to reduce overhead. While promising, these methods have not been explored in the context of long-horizon, multi-turn conversations, where cache reuse between different models becomes more fragile and error-prone. In our own observations, we found that sharing KV caches between differently fine-tuned models becomes increasingly problematic as the conversation grows longer, due to subtle drifts in behavior and context representation. Given the rapidly improving generalization ability of recent LLMs on diverse tasks, we argue that employing a single LLM is more \emph{robust} and \emph{cache-efficient} when managing instruction flow with keeping the conversation history across main and side tasks. 

To this end, we propose \myalg (\underline{\textbf{C}}ontextual \underline{\textbf{I}}nstruction \underline{\textbf{FL}}ow with \underline{\textbf{EX}}ecution), a novel execution strategy that extends the utility of KV cache from token-level reuse to task-level reuse. Rather than reloading the entire context for every sub-task, \myalg checkpoints the KV cache of the main conversation flow and reuses it by injecting only task-specific instructions into a side path. Sub-tasks are then executed efficiently on top of the preserved context without redundant processing. After execution, the model seamlessly rolls back to the main path using the cached state, avoiding the computational cost of full prefilled context. Additionally, we propose a hierarchical sub-task classification framework designed for smaller-scale LLMs. Instead of relying on multi-class selection, which is often unreliable for device-scale models—we perform a sequence of binary classifications in side paths, guided by our contextual instruction flow.

Since there have been no works considering sub-task execution in multi-turn conversations, we present long multi-turn conversation datasets (\eg TopiOCQA-Task+ and QReCC-Task+) by combining various task-specific datasets for conversational search, math problem solving, API calls, and casual chats. Then, through extensive experiments, we show that \myalg achieves significant computational savings while sacrificing less performance, offering a practical and scalable solution for multi-task dialogue systems in resource-constrained environments.

%% file: Method.tex
\begin{figure*}[t]
    \centering
    \includegraphics[width=0.9\linewidth]{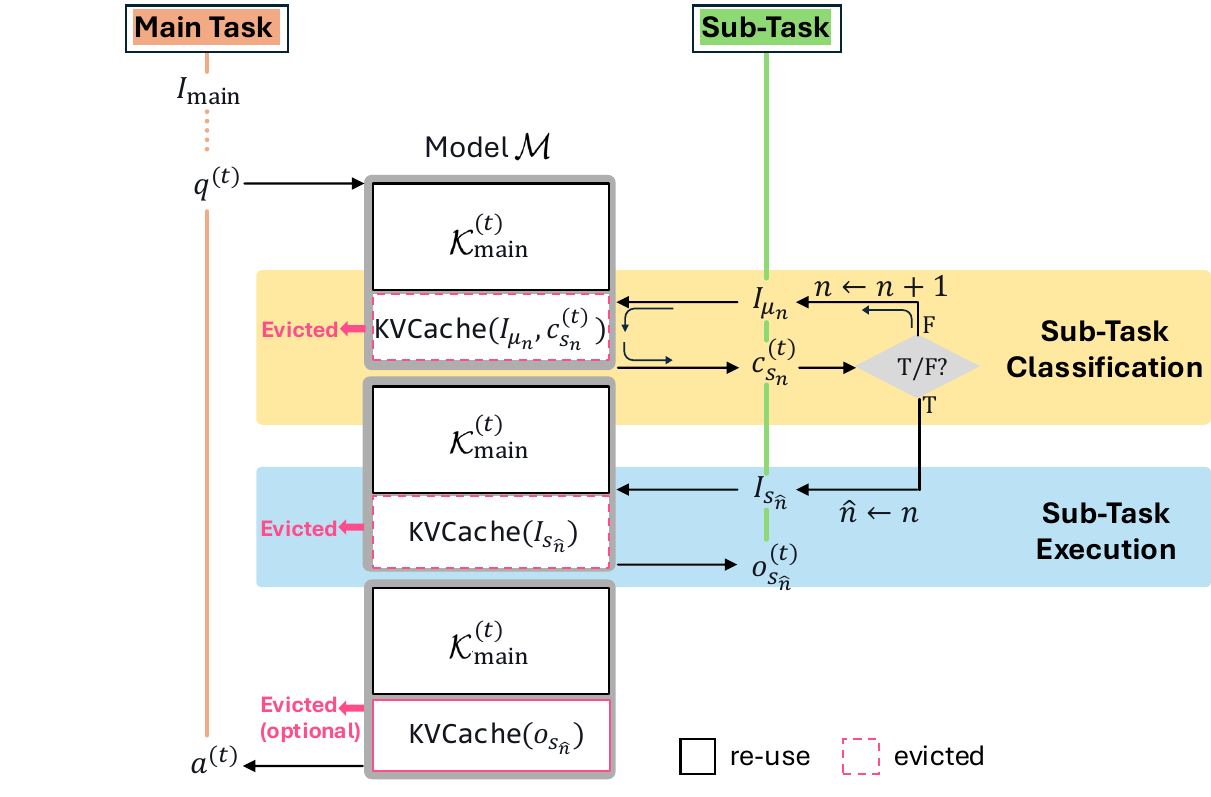}
    \caption{\textbf{Overall framework of the proposed \myalg.}}
    \label{fig:main}
\end{figure*}

\section{\myalg: \underline{\textbf{C}}ontextual \underline{\textbf{I}}nstruction \underline{\textbf{FL}}ow with \underline{\textbf{EX}}ecution}
\label{sec:method}

\subsection{Problem Formulation}
\label{sec:preliminary}




In a multi-turn interaction which is the main task $m$, an LLM $\mathcal{M}$ (here, we consider edge device-scaled model such as LLaMA3 3.1 8B) generates an answer $a^{(t)}$ for the given user query $q^{(t)}$ at each turn $t$. This main question and answer flow is represented by
\begin{equation}
\label{eq:mainflow} 
    a^{(t)} = m(\mathcal{M}, C_\mathrm{main}^{(t)}, q^{(t)}),
\end{equation}
where $C_\mathrm{main}^{(t)}$ denotes the conversational context of $m$ at turn $t$, consisting of the main task instruction $I_\mathrm{main}$ and all previous turns: 
\begin{equation}
\label{eq:mainflow_context}
\begin{array}{l}
C_\mathrm{main}^{(t)} = \\[1ex]
\quad
\begin{cases}
  \{ I_\mathrm{main} \} & \text{if } t = 1, \\
  \{ I_\mathrm{main},\ \{(q^{(t')}, r^{(t')}, a^{(t')})\}_{t'=1}^{t-1} \} & \text{if } t > 1,
\end{cases}
\end{array}
\end{equation}
where $r^{(t)}$ represents turn-specific context such as the external retrieval knowledge, which is optional but can lead to substantial increase $|C_\mathrm{main}^{(t)}|$.

To produce correct answer $a^{(t)}$, the LLM can optionally execute a proper auxiliary sub-task which supports the main answer but is not directly exposed to the user. The sub-tasks are designed to support $m$, and are represented as a set $\{s_1, \ldots, s_N\}$, where $s_n$ denotes the $n$th sub-task. Each sub-task produces an intermediate output $o_{s_n}^{(t)}$ based on its own context:
\begin{equation}
\label{eq:subflow} 
    o_{s_n}^{(t)} = s_n(\mathcal{M}, C_{s_n}^{(t)}, q^{(t)}),
\end{equation} 
where 
a sub-task-specific instruction $I_{s_n}$ is used in place of $I_{\mathrm{main}}$ in $C_{s_n}^{(t)}$. 

Also, before executing any sub-task $s_n$, a classification sub-task $\mu$ needs to be performed to determine which sub-task is necessary for the current turn. This classification task is also treated as another sub-task.

Then, in a single-LLM setting, whenever switching from the sub-tasks to the main task, the conversational context should be re-prefilled even though much of the context is redundant across tasks and turns. 
This repeated context reloading imposes significant computational overhead, which motivates our proposed approach for efficient sub-task execution.

\subsection{Methodology}
To enable efficient sub-task execution, we propose Contextual Instruction Flow with Execution dubbed \myalg. Fig.~\ref{fig:main} illustrates the overall flow. We also provide the set of prompt templates in Figs.~\ref{fig:inf_main},~\ref{fig:inf_cls},~\ref{fig:inf_sub}.

\noindent\textbf{Sub-Task Execution \& Rollback.}  Once a user query $q_t$ is issued, we create a \textit{checkpoint} at that moment—preserving the key-value (KV) cache of the entire main path up to $q_t$. For instance, when $t=1$
\begin{equation}
\label{eq:kv_main}\mathcal{K}_{\text{main}}^{(1)} = \texttt{KVCache} \left(  C_{\mathrm{main}}^{(1)}, q^{(1)} \right).
\end{equation}

Based on this cached state, we branch into the appropriate sub-task path. Instead of reprocessing the full conversation context, we prefill only the sub-task instruction $I_{s_n}$ on top of the preserved KV cache. 
\begin{equation}
\label{eq:subtask_forward}
o_{s_{n}}^{(1)} = s_n ( \mathcal{M}, I_{s_n};\ \mathcal{K}_{\text{main}}^{(1)})
\end{equation}

The sub-task $s_n$ is then executed efficiently. Empirically, we observe that although the main instruction $I_{\mathrm{main}}$ is implicitly retained via the KV cache, the sub-task reliably adheres to $I_{s_n}$, without interference from $I_{\mathrm{main}}$.

After completing the sub-task, we evict the KV cache related to the sub-task instruction (e.g., $I_{s_n}$) or complex reasoning process and, if necessary, retain the cache associated with the sub-task's final output $o_{s_{n}}^{(1)}$. 
\begin{equation}
\label{eq:rollback}
\mathcal{K}_{\text{rollback}}^{(1)} = \mathcal{K}_{\text{main}}^{(1)} \oplus 
\texttt{KVCache} \left( o_{s_{n}}^{(1)} \right)
\end{equation}
where $\oplus$ denotes concatenation operation. We can then \emph{roll back} to the main path and the model to continue the main task without re-prefilling the full conversational history. 
\begin{equation}
\label{eq:mainflow} 
    a^{(1)} = m(\mathcal{M};\mathcal{K}_{\text{rollback}}^{(1)} )).
\end{equation}
Here, $r^{(1)}$ can be prefilled, if necessary, i.e. $a^{(1)} = m(\mathcal{M}, r^{(1)};\mathcal{K}_{\text{rollback}}^{(1)} ))$.

Notice that, when $t>1$, 
the KV cache is shared across turns as well, then it is simply updated after the model outputs $a^{(t)}$:
\begin{equation}
\label{eq:kv_main}
\mathcal{K}_{\text{main}}^{(t+1)} = \mathcal{K}_{\text{rollback}}^{(t)} \oplus \mathcal{K}_{a}^{(t)} \oplus \texttt{KVCache} \left(  q^{(t+1)} \right)
\end{equation}
where 
$\mathcal{K}_{a}^{(t)}$ is the added KV cache during answering in the main task. Therefore, only the KV cache for $q^{(t+1)}$ is computed in prefill.

This structured mechanism is what we refer to as \textit{contextual instruction flow}, where the shared conversational context is reused, and only task-specific instructions are attached or evicted to enable efficient task switching. This contextual instruction flow allows substantially reducing the overall computational cost of switching the main and sub-tasks during multi-turn conversations with a single LLM.



\begin{table}[t]
  \centering
  \begin{adjustbox}{width=0.85\linewidth}
  \begin{tabular}{@{}lcccc@{}}
    \toprule
    Model & \makecell{LLaMA\\(8B)} & \makecell{Mistral\\(7B)} & \makecell{Qwen\\(7B)} & \makecell{GPT-4\\(>100B)}\\
    \midrule
    Acc (\%) & 33.93 & 44.31 & 42.13 & 98.21 \\
    \bottomrule
  \end{tabular}
  \end{adjustbox}
  \caption{\textbf{Challenge of multi-choice sub-task classification set-up on smaller-scaled models.} Accuracy (\%) is reported on the dataset extended from TopiOCQA.}
  \label{tab:method_cls}
\end{table}

\noindent\textbf{Sub-Task Classification.}  
As described in Section~\ref{sec:preliminary}, the objective of sub-task classification is to determine which sub-task should be executed. In our work, it is also formulated using the proposed contextual instruction flow strategy. 

As shown in Table~\ref{tab:method_cls}, small-scale language models struggle to comprehend the definitions of multiple sub-tasks simultaneously, failing to make accurate selections under a multi-choice setup. To address this limitation, we propose a hierarchical selection strategy.

Specifically, we first assign priorities to the available sub-tasks using a larger LLM, offline. Then, at runtime, the on-device model performs binary classification per sub-task in descending order of priority, varying only the class-specific instruction in the contextual instruction. 
\begin{equation}
\label{eq:subtask_forward}
c_{s_n}^{(t)} = \mu_n ( \mathcal{M}, I_{\mu_n};\ \mathcal{K}_{\text{main}}^{(t)})
\end{equation}

Then, if a sub-task with higher priority is classified as necessary, the hierarchical process is terminated, and the selected sub-task $s_{\hat{n}}$ is executed.

It is important to note that these classification decisions are also made using the same single LLM instance. As a result, executing multiple binary classifications can lead to a significant increase in prefilling cost. However, thanks to our contextual instruction flow design, only the task-specific classification instructions need to be prefetched, which minimizes redundant computation and ensures efficient task routing.

%% file: Exp.tex
\section{Multi-turn Dataset Construction}
\label{sec:dataset}
\vspace{-0.2cm}
Following~\cite{yi2024survey}, we consider broadly two kinds of conversations: task-oriented conversation (conversational search, math problem, multi-modal question) and open-domain conversation (casual chat). Then, sub-tasks are mapped as follows:

\paragraph{(1) Conversational Search - Query Rewriting.}  
We employ TopiOCQA~\cite{TopiOCQA} and QReCC~\cite{QReCC}, both of which consist of multi-turn conversational queries. These datasets contain context-dependent questions involving pronouns, ellipses, and coreference, requiring disambiguation based on prior conversation turns. Hence, in the sub-task, LLM clarifies the original question into a standalone question. Then, an off-the-shelf retriever (e.g., BGE-Large~\cite{BGE}) retrieves evidence passages which is given to the main task.


\paragraph{(2) Math Problem - Reasoning \& Answer.}  
To this end, we incorporate the GSM8K dataset~\cite{cobbe2021gsm8k}, which provides both problem statements and detailed reasoning steps. We randomly sample math problems and concatenate them naturally into existing TopiOCQA and QReCC, respectively. In the side-path execution flow, the LLM is in-context learned, and outputs the reasoning and answer are generated. Then, only the final answer is returned to the main path. This design helps reduce the contextual burden on the main flow while enabling intermediate computation in the sub-task flow.

\paragraph{(3) Multi-Modal - API Call.}  
Here, we exploit the Gorilla dataset~\cite{patil2023gorilla}, which contains API specifications and their expected invocation patterns. Based on a randomly selected API, then a larger LLM (e.g., LLaMA3.1 70B~\cite{dubey2024llama}, GPT4~\cite{achiam2023gpt}) generates a multi-modal question like \textit{“What is the color of the whale in the Pacific Ocean [Image]?”}. In sub-task, LLM is instructed to select a proper API call. 
Then, the textual description for the multi-modal input from an off-the-shelf model is returned to the main task.

\paragraph{(4) On-going Casual Chat - No Sub-Task.}  
In practice, not every user query requires a sub-task. To reflect such cases, we include examples where the user is in the middle of casual conversation with the assistant and no sub-task execution is necessary. Specifically, we append several turns of casual, non-task-oriented turns—covering topics such as weather, hobbies, or daily life—on top of task-oriented interactions from categories (1)–(3). These casual turns are generated using a larger LLM. These cases serve as negative examples for sub-task classification and help balance the task distribution.

\paragraph{(5) Last Casual Chat - Chat Summary.}  
On edge devices, AI assistants are often personalized to the user, making it crucial to understand user preferences and behavior across interactions. To support this, we introduce a summarization sub-task that captures the essence of a casual conversation once it concludes. Namely, in the sub-task, LLM is instructed to summarize the user's behavior or interests based on the conversation. This summary may be saved in an external knowledge hub.

\paragraph{} 

\noindent{- Commonly, to ensure a coherent multi-turn flow and maintain topic consistency, we include the full conversation history from TopiOCQA or QReCC within the prompt when generating additional casual chat or API-related turns using large LLMs (the corresponding prompts are included in Appendix). This allows the generated content to remain contextually grounded and avoids unnatural topic shifts across turns.
Finally, since we use two different conversational search datasets, the resulting multi-turn, multi-task datasets are dubbed \textbf{TopiOCQA-Task+} and \textbf{QReCC-Task+}, respectively, which enable us to evaluate our proposed execution framework (\myalg) in a unified yet diverse conversational setting. We provide more details in Appendix~\ref{app:dataset}. 

\section{Experimental Results}
\vspace{-0.2cm}

\subsection{Baselines}
\label{sec:baseline}
To show the effectiveness of our method, we compare the proposed \myalg with three baselines:

-\textit{Full Re-load}: We assume a single LLM is runnable due to the resource constraints of edge devices. Hence, in this baseline, all the conversation history is re-loaded (pre-filled as textual inputs) whenever main or sub-tasks are switched without considering KV cache reuse across tasks. 

-\textit{Recent Re-load}: For less computational burden of re-loading the entire conversation history, only the recent several turns are retained, truncating the earlier turns. Here, the recent five turns are used.

-\textit{Seamless}: To mitigate the burden of the re-loading, we can execute all the tasks in a single flow without task switching. Here, all the sub-task execution instructions and results are included in the main task seamlessly.

-\textit{Chain-of-Model (CoM)} ~\cite{lyu2025streamliningcollaborativechainmodels}: In this baseline, we train the sub-task-specific LoRA adapters~\cite{hu2021lora} starting from the same pre-trained model, and CoM employs learnable prompt tokens to increase adaptability across the differently-trained models in sharing the KV-caches of common parts. 

\subsection{Evaluation Metrics}

\noindent\textbf{Sub-Task Classification.}  
To evaluate the effectiveness of the proposed per-task binary classification, we measure the accuracy of the classification results. Each sub-task is independently predicted in its own path, and overall classification performance is reported as the average accuracy across tasks.

\noindent\textbf{Sub-Task Execution.}  
For the query rewriting sub-task, we use the rewritten query to retrieve passages via an off-the-shelf retriever, BGE-Large~\cite{BGE}. Since both TopiOCQA and QReCC provide ground-truth retrieval annotations, we report standard information retrieval metrics: nDCG@3 reflects relevance and order of top-3 results, and Hit@k indicates whether any relevant passage is retrieved within the top-$k$ passages where we set $k=5$.
Since the chat summary and API call have no static, fixed ground truth, we utilize the GPT-eval metric, the coherence~\cite{zhong2023memorybankenhancinglargelanguage}. This means the side-task output is evaluated if it is aligned logically and contextually with the corresponding side-task instruction. The detailed metric prompt is included in Appendix. 
For the math problem, the sub-task output is directly used in the main task; therefore, we assess it as part of the main task's answer.

\noindent\textbf{Main-Task Execution.}  
We utilize the GPT-eval metrics, coherence and correctness, based on~\cite{zhong2023memorybankenhancinglargelanguage}. Correctness metric is used only for the task with the static, fixed ground-truth, i.e. math problem.


\subsection{Target models} 
For the edge-device model, we employ LLaMA3.1-8B-Inst~\cite{dubey2024llama}, Mistral-v0.2-Inst 7B~\cite{jiang2023mistral}, Qwen2.5-Inst-7B~\cite{yang2024qwen2} models. 


\begin{table*}[t]
  \centering
  \begin{adjustbox}{width=\linewidth}
  \begin{tabular}{@{}llccc|c@{}}
    \toprule
    \multirow{3}{*}{Dataset} & \multirow{3}{*}{Method} & \multicolumn{3}{c}{On-device-scale model} & Server-scale model \\
    \cmidrule{3-5} \cmidrule{6-6} 
    && LLaMA3.1-Instruct & Mistral-Instruct-v0.2 & Qwen2.5-Instruct & GPT-4\\
    && (8B) & (7B) & (8B) & (>100B)\\
    \midrule
    \multirow{2}{*}{TopiOCQA-Task+} & Multi-choice & 33.93 & 44.31 & 42.13 & 98.21\\
    & \textbf{Per-task binary} & \textbf{88.01} & \textbf{83.68} & \textbf{84.4} & 98.55\\
    \bottomrule
    \multirow{2}{*}{QReCC-Task+} & Multi-choice & 51.83 & 57.66 & 59.06 & 96.52\\
    & \textbf{Per-task binary} & \textbf{82.04} & \textbf{80.68} & \textbf{83.97} & 97.31\\
    \bottomrule
  \end{tabular}
  \end{adjustbox}
  \caption{\textbf{Sub-task classification results.} Accuracy (\%) are reported on multi-choice strategy and the proposed per-task binary classification varying models on TopiOCQA-Task+ and QReCC-Task+. The best results are in bold.}
  \label{tab:exp_cls}
\end{table*}

\begin{figure*}[t]
    \centering
    \begin{subfigure}{0.6\linewidth}\includegraphics[width=\linewidth]{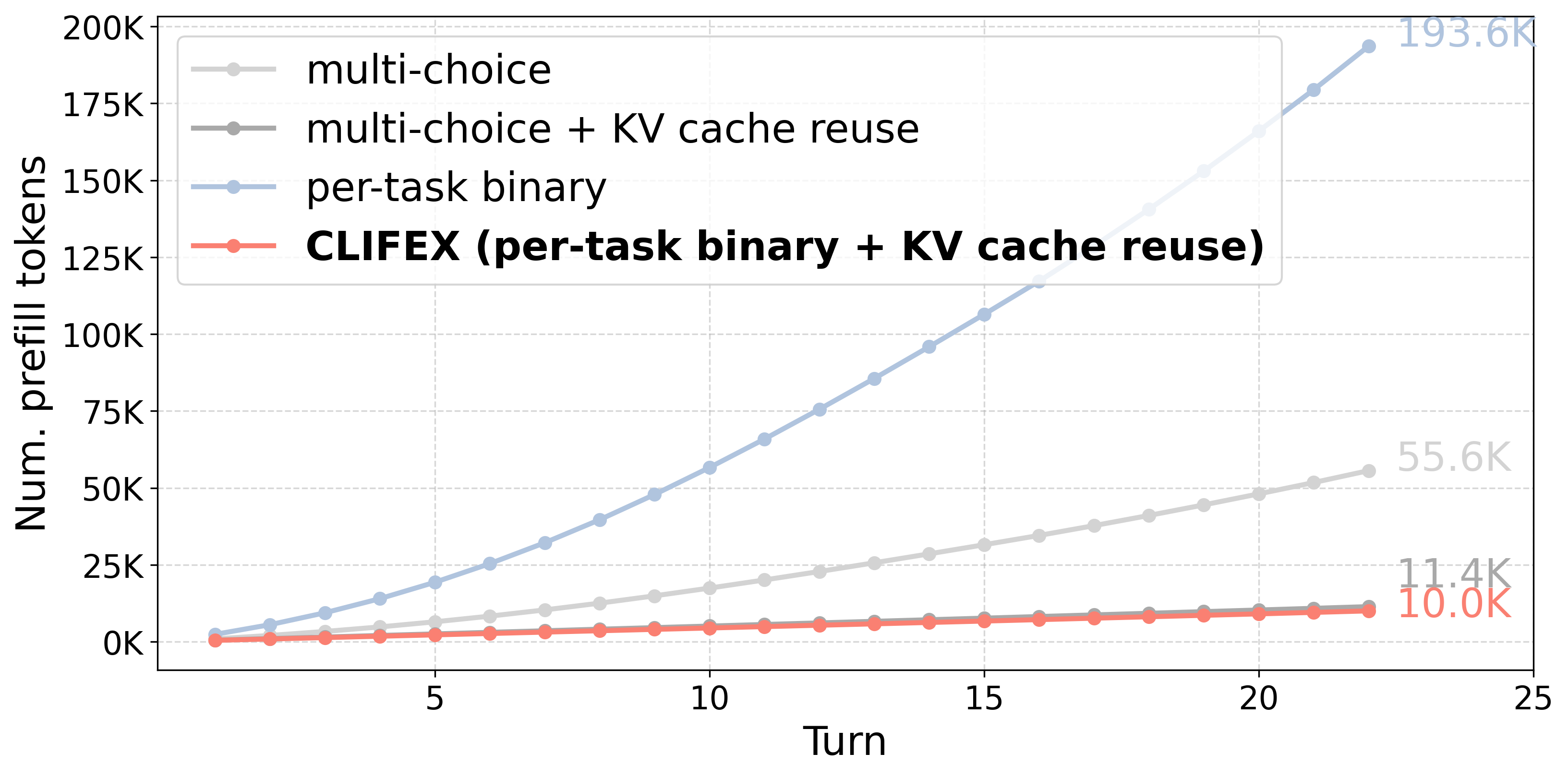}
    \end{subfigure}
    \vspace{-0.2cm}
    \caption{
    \textbf{Total prefilled tokens for sub-task classification until each turn} in LLaMA3.1-Instruct (8B) on  TopiOCA-Task+.}  
    
    \label{fig:exp_cls_tok}
    \vspace{-0.2cm}
\end{figure*}

\subsection{Results}

\noindent\textbf{Sub-Task Classification.}  
As described in Sec.~\ref{sec:dataset}, each user query is classified into one of five sub-task categories: query rewriting, math solving (reasoning \& answer), API call, chat summary, and no sub-task. Following our per-task binary classification with task priority strategy, we assign a fixed priority order to the five classes based on their class-wise recall rates measured by a larger oracle model. Classes with clearer semantic boundaries receive higher priority, while more ambiguous classes are ranked lower. Specifically, the priority order is: API call $\rightarrow$ Math solving $\rightarrow$ Query rewriting $\rightarrow$ Summarization $\rightarrow$ None. In cases where multiple sub-tasks are simultaneously detected, the one with higher priority is selected (e.g., API call over query rewriting). As shown in Table~\ref{tab:exp_cls}, for all the models and datasets, the proposed approach significantly outperforms the multi-choice strategy, which often struggles to detect the correct sub-task. 

Also, in terms of the efficiency, Fig.~\ref{fig:exp_cls_tok} plots the required total number of pre-fill tokens up to a given turn. Although the per-task classification is effective, it requires an extremely high number of tokens (193.6K for 22 turns) without applying the proposed \myalg. With \myalg, per-task binary (orange) even uses slightly less pre-fill tokens than multi-choice (dark gray), since the instructions of sub-tasks are designed to include shared parts (see the Appendix). Hence, our per-task classification with \myalg is highly effective and efficient. 


\begin{table*}[t]
  \centering
  \begin{adjustbox}{width=\linewidth}
  \begin{tabular}{@{}cclccccc@{}}
    \toprule
    & & & \multicolumn{2}{c}{Sub-Task} & & Main Task \\
    \cmidrule{4-5} \cmidrule{7-7}
    && & Query Rewrite &  Chat Summary & & Answering \\
    Dataset&Model& Method & (ACC, DCG)) & (Cohere) & & (Cohere, Correct) \\
    \midrule
    \multirow{15}{*}{TopiOCQA-Task+} &\multirow{5}{*}{\makecell{LLaMA3.1-Instruct\\(8B)}} 
    & Full Re-load  & (63.04, 31.07) & 98.77 && (86.45, \textbf{56.50}) \\
    && Recent Re-load  & (62.53, 30.83) & 98.65 &&  (85.90, 56.20) \\
    && Seamless  & (57.13, 21.34) & 97.85 &&  (83.30, 54.10)  \\
    && CoM~\cite{lyu2025streamliningcollaborativechainmodels}  & (61.41, 29.82) & 98.55 &&  (85.40, 55.20)  \\
    && \textbf{Proposed}   & (\textbf{63.04}, \textbf{31.34}) & \textbf{98.92} && (\textbf{87.57}, 55.59) \\
    \cmidrule{2-7}
    &\multirow{5}{*}{\makecell{Mistral-Instruct-v0.2\\(8B)}}
    & Full Re-load  & (61.10, \textbf{29.84}) & 99.46 && (82.26, \textbf{54.30}) \\
    && Recent Re-load   & (59.42,27.81) & 99.30 && (81.10, 53.60) \\
    && Seamless  & (53.13, 22.45) & 98.50 && (78.90, 52.40) \\
    && CoM~\cite{lyu2025streamliningcollaborativechainmodels}  & (59.91,29.31) & 99.15 && (81.00, 53.25) \\
    && \textbf{Proposed}   & (\textbf{61.45}, 29.83) & \textbf{99.75} && (\textbf{83.75}, 52.75) \\
    \cmidrule{2-7}
    &\multirow{5}{*}{\makecell{Qwen2.5-Instruct\\(8B)}}
    & Full Re-load  & (\textbf{64.29}, \textbf{32.18}) & \textbf{99.75} && (\textbf{75.86}, \textbf{57.11}) \\
    && Recent Re-load   & (63.70, 31.85) & 99.65 && (75.50, 56.80) \\
    && Seamless  & (58.20, 29.60) & 98.85 && (73.10, 54.90) \\
    && CoM~\cite{lyu2025streamliningcollaborativechainmodels}  & (62.90, 31.30) & 99.40 && (74.50, 55.80) \\
    && \textbf{Proposed}   & (63.12, 31.59) & 99.51 && (74.18, 55.57) \\
    \midrule
    \multirow{15}{*}{QReCC-Task+}&\multirow{5}{*}{\makecell{LLaMA3.1-Instruct\\(8B)}} 
    & Full Re-load  & (77.37, \textbf{28.12}) & \textbf{97.97} & & (74.81, \textbf{65.45}) \\
    && Recent Re-load  & (76.70, 27.70) & 97.51 & & (73.88, 64.32) \\
    && Seamless  & (71.50, 24.90) & 96.11 & & (71.04, 60.12) \\
    && CoM~\cite{lyu2025streamliningcollaborativechainmodels}   & (75.90, 27.30) & 97.04 & & (74.43, 64.10) \\
    && \textbf{Proposed}   & (\textbf{77.43}, 27.59) & 97.74 & & (\textbf{75.32}, 65.04) \\
    \cmidrule{2-7}
    &\multirow{5}{*}{\makecell{Mistral-Instruct-v0.2\\(8B)}}
    & Full Re-load  & (74.91, 46.72) & \textbf{97.14} & & (66.94, 57.38) \\
    && Recent Re-load  & (73.80, 45.10) & 96.04 & & (65.82, 55.76) \\
    && Seamless  & (68.10, 41.00) & 95.53 & & (62.11, 52.71) \\
    && CoM~\cite{lyu2025streamliningcollaborativechainmodels}   & (73.90, 44.90) & 96.22 & & (65.32, 55.05) \\
    && \textbf{Proposed}   & (\textbf{76.04}, \textbf{48.48}) & 96.99 & & (\textbf{67.08}, \textbf{57.95}) \\
    \cmidrule{2-7}
    &\multirow{5}{*}{\makecell{Qwen2.5-Instruct\\(8B)}}
    & Full Re-load  & (\textbf{77.08}, \textbf{49.26}) & \textbf{99.72} & & (\textbf{80.17}, \textbf{70.55}) \\
    && Recent Re-load  & (75.90, 47.90) & 98.55 & & (77.31, 68.00) \\
    && Seamless  & (69.80, 43.20) & 96.42 & & (74.13, 64.19) \\
    && CoM~\cite{lyu2025streamliningcollaborativechainmodels}   & (74.80, 46.60) & 98.60 & & (78.51, 68.72) \\
    && \textbf{Proposed}   & (76.04, 48.48) & 99.68 & & (79.57, 69.12) \\
    \bottomrule
  \end{tabular}
  \end{adjustbox}
  \caption{\textbf{Evaluation results (\%) for sub-task \& main-task execution} on TopiOCQA-Task+ and QReCC-Task+. Sub-tasks of API call and math problem are evaluated as answering quality (correctness) in the main task.}
  \label{tab:exp_acc1}
\end{table*}

\begin{figure*}[t]
    \centering
    \begin{subfigure}{0.485\linewidth}\includegraphics[width=\linewidth]{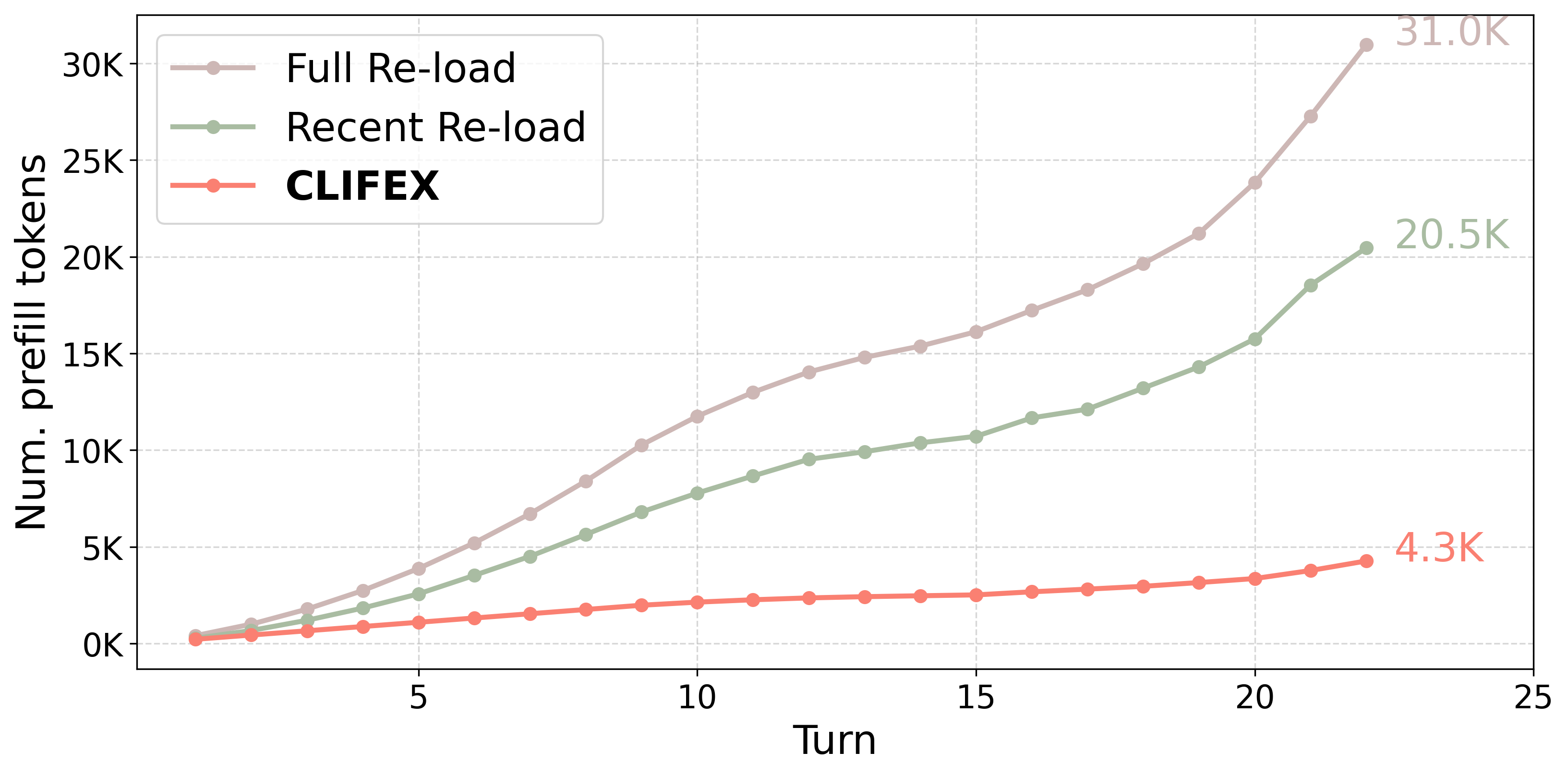}
    \caption{Sub-Task}
    \end{subfigure}
    \begin{subfigure}{0.485\linewidth}\includegraphics[width=\linewidth]{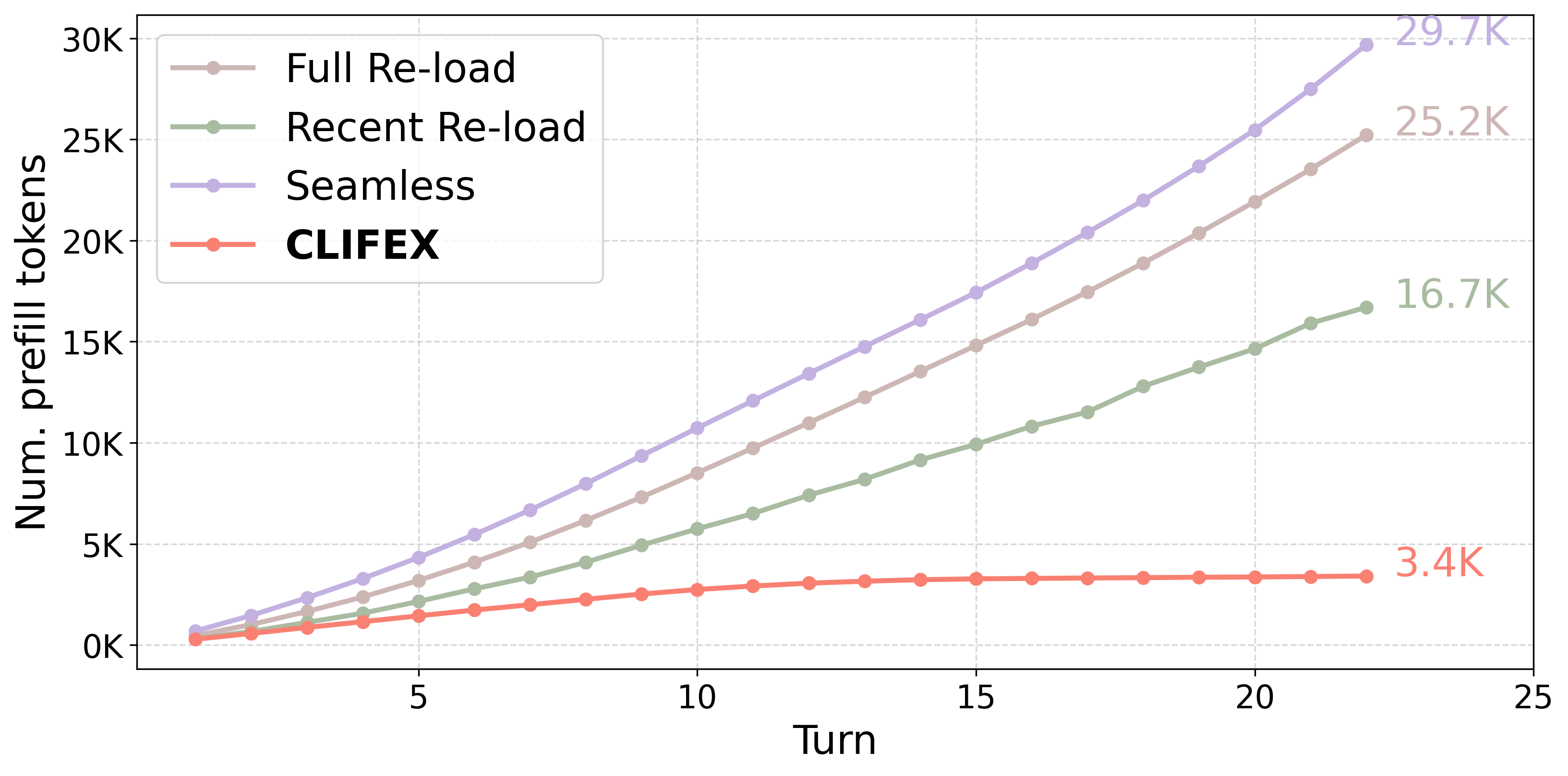}
    \caption{Main-Task}
    \end{subfigure}
    \caption{
    \textbf{Turn-wise prefilled tokens for main-task and sub-task execution until each turns} in LLaMA3.1-Instruct (8B) on TopiOCA-Task+.} 
    \label{fig:exp_acc_tok2}
    \vspace{-0.3cm}
\end{figure*}

\begin{table}[t]
  \centering
  \begin{adjustbox}{width=\linewidth}
  \begin{tabular}{@{}llccccc@{}}
  \toprule
  \multirow{2}{*}{Method} & \multirow{2}{*}{Stage} & \multicolumn{5}{c}{Turn} \\
  \cmidrule{3-7}
  & & 2 & 5 & 10 &15& Last \\
  \midrule
  \multirow{2}{*}{Full Re-load} & Prefill & 4.35 & 25.38 & 86.90 & 165.76 & 331.76 \\
  & Generation & 1.94 & 8.27 & 18.47 & 25.13& 39.43\\
  \midrule
  \multirow{3}{*}{\textbf{Proposed}} & Prefill & 1.26 & 5.06 & 11.30& 15.88 & 23.51 \\
  & Prefill (\textit{bp}) & \textbf{0.95} & \textbf{3.84} & \textbf{8.56} & \textbf{11.60} & \textbf{16.79} \\
  & Generation & 2.02 & 8.34& 18.80& 25.52& 41.15\\
  \bottomrule
  \end{tabular}
  \end{adjustbox}
  \caption{\textbf{Latency ($\downarrow$) in terms of seconds on edge device.} `\textit{bp}' denotes the batch processing in sub-task classification.}
  \label{tab:latency}
\end{table}

\noindent\textbf{Sub \& Main-Task Execution.}  
We compare the proposed \myalg against the four baselines introduced in Sec.~\ref{sec:baseline}. As shown in Table~\ref{tab:exp_acc1}, the Seamless baseline exhibits poor performance across both sub-tasks and the main task. This degradation indicates that accumulated main \& sub-task execution history interfere with LLM's contextual understanding ability,highlighting the necessity of independent sub-task execution in separate flows.

The Full Re-load baseline yields reliable performance owing to prefilling clean task-specific instructions, but at the significant cost of context re-loading overhead, as in Fig.~\ref{fig:exp_acc_tok2}. Compared to the Full Re-load, the Recent Re-load baseline reduces prefilling cost by reloading only the truncated recent turns; however, it still requires much more prefilling tokens than our \myalg. Moreover, it suffers from performance degradation. Although the CoM~\cite{lyu2025streamliningcollaborativechainmodels} learns models to increase the adaptability of KV caches across different models, its performance is lower than the single model inference (ours). We can see that cache-sharing across heterogeneous models becomes fragile and error-prone as the number of turns increases.

In contrast, the proposed \myalg achieves comparable or superior performance to the Full Re-load, while incurring significantly less prefilling cost, as in the orange of Fig.~\ref{fig:exp_acc_tok2}. This demonstrates that \myalg strikes an effective balance in terms of both accuracy and efficiency in multi-turn, multi-task execution under a single LLM. More analyses are in Appendix.

\noindent \textbf{On-Device Latency}: Table~\ref{tab:latency} shows the on-device latency of prefilling and generation. We measured this latency on the mobile edge device equipped with Snapdragon\textsuperscript{\textregistered}~8 Elite chipset. Here, prefilling latency encompasses the latencies of sub-task classification, sub-task execution, and main task execution for each turn. Similarly, generation latency is computed. Technically, the generation latency is about 50 times that of the prefilling latency. Nevertheless, in the baseline, the prefilling latency drastically increases beyond generation latency as the number of turns grows. This underscores the importance of an efficient prefilling strategy in multi-turn interactions. Our method achieves very low latency, which is even lower than typical generation latency, making it suitable for practical use. Also, as the per-task classification can be batch-processed, then the latency is further reduced where the classification latency was reduced by half compared to the case without batch processing.

%% file: Related.tex
\section{Related Works}
\vspace{-0.2cm}

\noindent{\textbf{Efficient LLM Inference}.} To reduce the computational and memory overhead in inference, several works have explored various strategies addressing KV caches. Approaches targeting task-agnostic inference efficiency are based on a single LLM. Then, they enhance KV cache efficiency via strategies such as sharing KV caches across layers~\cite{yang2024kvsharerefficientinferencelayerwise}, computing KV caches for a subset of layers~\cite{wu2024layer}, and saving KV caches for fast GPU-to-CPU offloading~\cite{lee2025efficient}.

In contrast, methods assuming different tasks utilize two or more task-specific models. They leverage fine-tuning or model adaptation to facilitate KV cache sharing and reuse. FTHSS~\cite{lyu2025streamlining} exploited learnable prompt tokens for KV cache sharing in LLMs. Similarly, DroidSpeak~\cite{liu2024droidspeak} optimizes context sharing in fine-tuned LLMs by selectively recomputing critical layers while reusing non-critical KV cache segments. KVLINK~\cite{yang2025kvlink} pre-computes and reuses KV caches for document segments, employing mixed-data fine-tuning. While efficient, these methods require model modifications, often fine-tuning. Their applicability in multi-turn interactions, however, remains underexplored.

\noindent{\textbf{Multi-turn Conversation.}} In LLM, the multi-turn conversation can be categorized into task-oriented dialogue (TOD) and open-domain dialogue (ODD)~\cite{yi2024survey}. In TOD, LLMs assist users in achieving specific goals, often requiring structured interaction and the potential management of external APIs or retrieval of external knowledge if necessary. ODD focuses on engaging in free-form, natural conversations without a predefined task constraint. 

While these distinctions exist, LLMs exhibit substantial performance drop in multi-turn conversations~\cite{laban2025llms}, which stems from the inherent difficulty for LLMs in effectively managing the dialogue flow over turns. 
Effectively handling multi-turn conversations necessitates sophisticated capabilities~\cite{laban2025llms}—understanding history, interpreting past exchanges, and adapting to evolving user objectives—which introduce complexities like maintaining coherence, ensuring alignment with shifting intentions, and mitigating cumulative errors and contextual drift. Mixed TOD/ODD dialogues present further complexities,
Such interactions necessitate capabilities like dynamic task classification, instruction adaptation based on the identified task, and potentially sub-task execution for TOD. Nevertheless, comprehensive datasets designed to capture the nuances of these complex multi-turn interactions have not yet curated. In this work, we carefully designed datasets for assessing LLMs on such complex multi-turn scenarios.


%% file: appendix.tex
\appendix
\supptitle
\vspace{1.0cm}

\section{Challenge of multi-choice classiciation set-up}
In contrast with general multi-choice questions where each choice option includes short and few words, a certain amount of task definition/explanation and the task name are involved in the choice options of our benchmark. This is the task description that we used in our sub-task classification (We use the same description in our per-task binary and the multi-choice baseline for fairness):

\noindent(A) Query rewrite – \textit{Select this option if the target user question contains any context-dependent words (e.g., "this," "that," "it," "these", "the xxxx") that rely on previous turns to be fully understood. Even if the query is grammatically correct, it must be rewritten for clarity when its meaning is ambiguous without prior context. }

\noindent(B) API Call – \textit{Select this option if the response to the target user question can be made by performing an API Call of image, audio, speech models.}

\noindent(C) Finish several turns of casual chat – \textit{Select this option if the conversation included multiple turns of casual chat or small talk, and in the target user question intends to conclude this talk.}

\noindent(D) Math Problem - \textit{Select this option if the last user interaction involves solving a math problem or requires mathematical computation.}

\noindent(E)  Clear user interaction and casual chat – \textit{Select this option if the intent of the target user question is keeping casual chat with AI Assistant with no needs of (A), (B), (C). e.g. "Hi", "I'm tired". In specific, the target user question is stand-alone understandable without previous turns. The target user question does not need API call of image, audio, speech models. The intent of the target user question does not conclusion of the conversation.}

We believe that 7B/8B models struggle to fully understand the definition of different tasks and the criteria categorizing them, at once. Whereas, larger models (e.g. GPT4) inherent capability is further stronger than those models. Hence, in our benchmark, these models revealed limitation under the multi-choice set-up. The proposed per-task binary classification is a practical solution, and it efficiency is resolved with the propose cache reusing strategy.

\section{Without $r^{(t)}$}
In conversational search (RAG), retrieved passages $r^{(t)}$ often constitute a large portion of the conversation history. While reducing prefill tokens by omitting passages at each turn (maintaining only the full Q\&A history) is possible, we observed a severe degradation in retrieval performance: ACC and DCG scores dropped from 63.04/31.34 to 50.12/20.70.

\section{Token Cost Analysis on QReCC-Task+}
\begin{figure}[h]
    \centering
    \begin{subfigure}{\linewidth}\includegraphics[width=\linewidth]{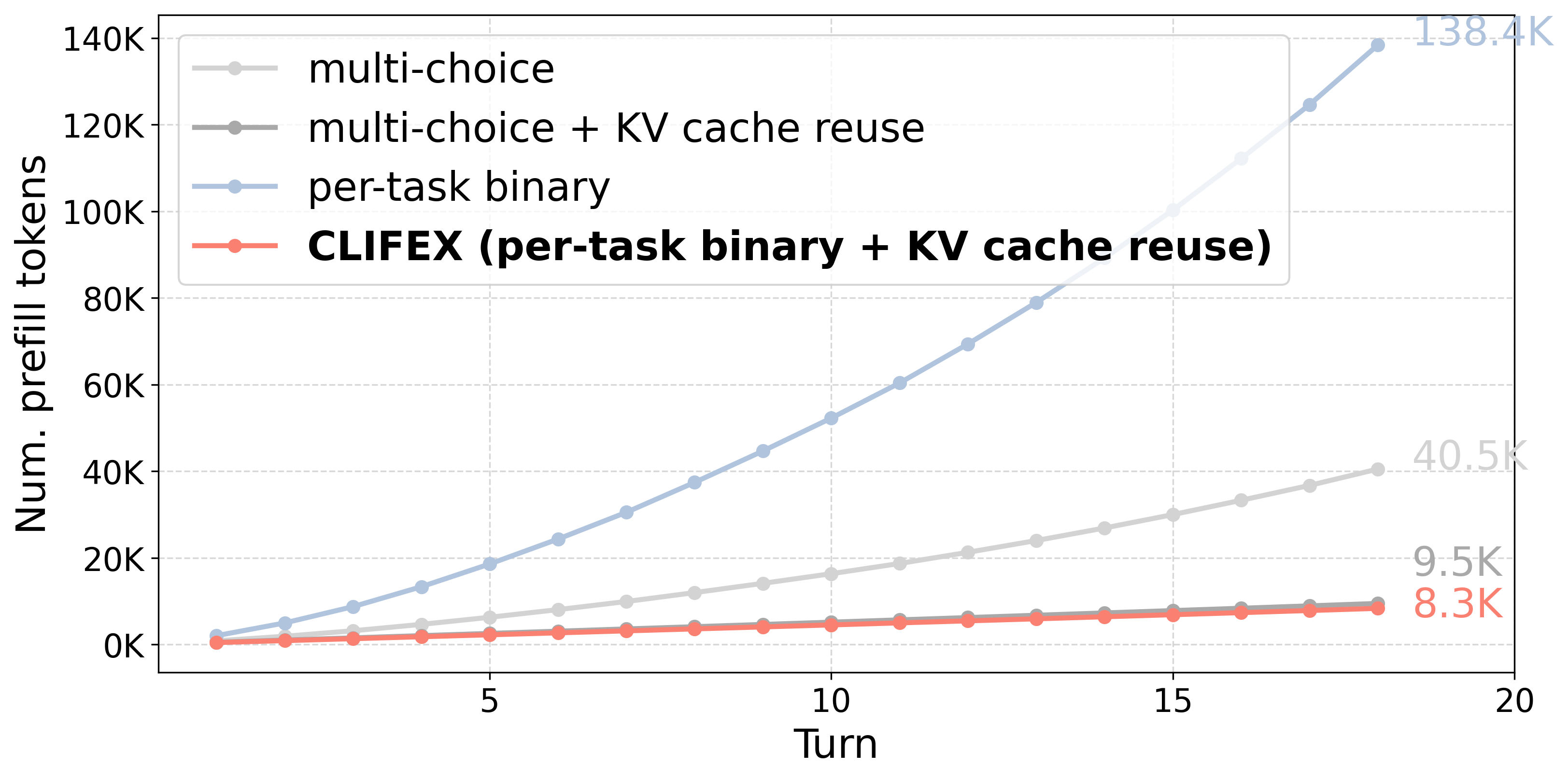}
    \caption{Sub-task Classification}
    \end{subfigure} \\
    \begin{subfigure}{\linewidth}\includegraphics[width=\linewidth]{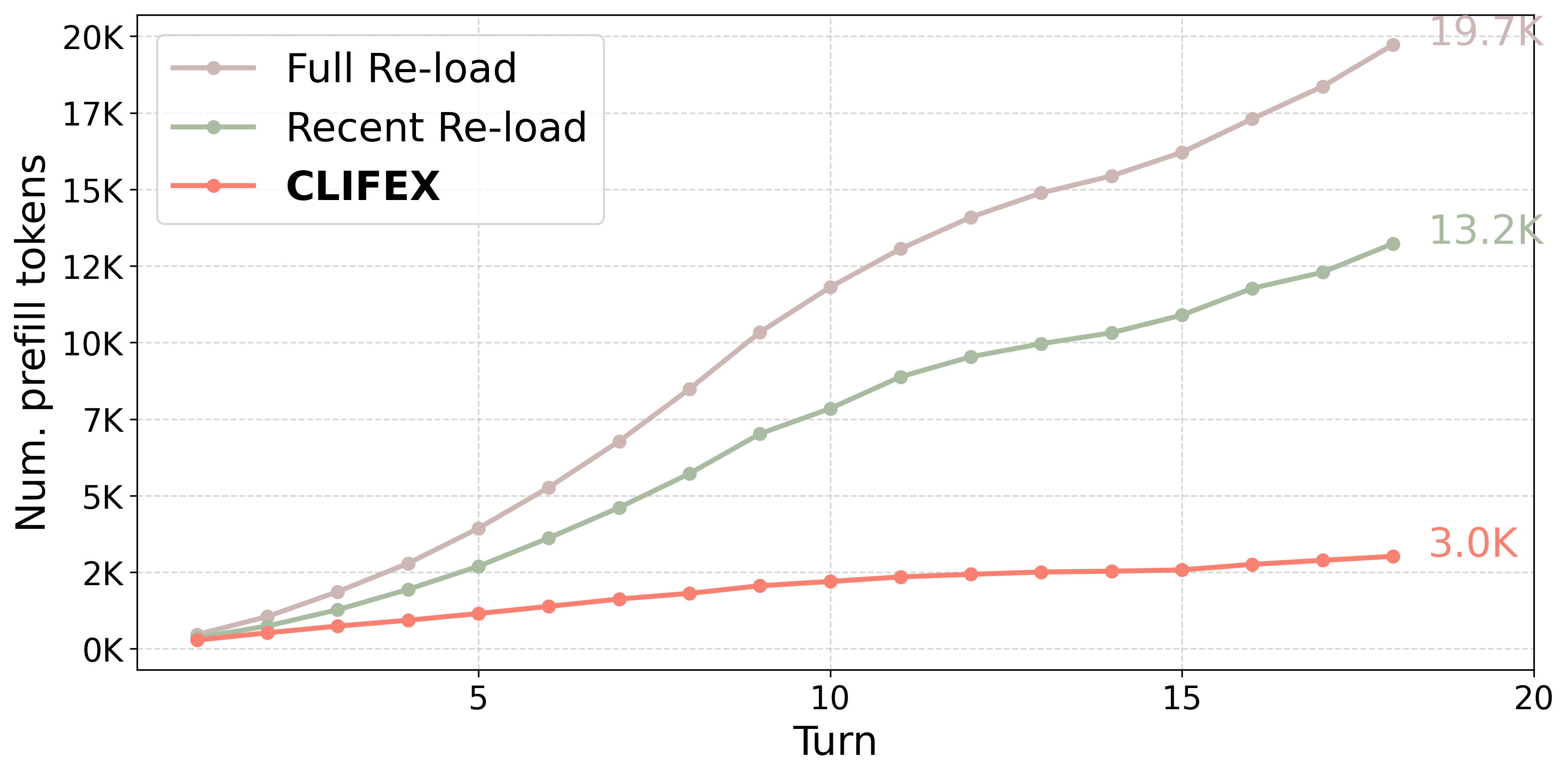}
    \caption{Sub-task execution} 
    \end{subfigure} \\
    \begin{subfigure}{\linewidth}\includegraphics[width=\linewidth]{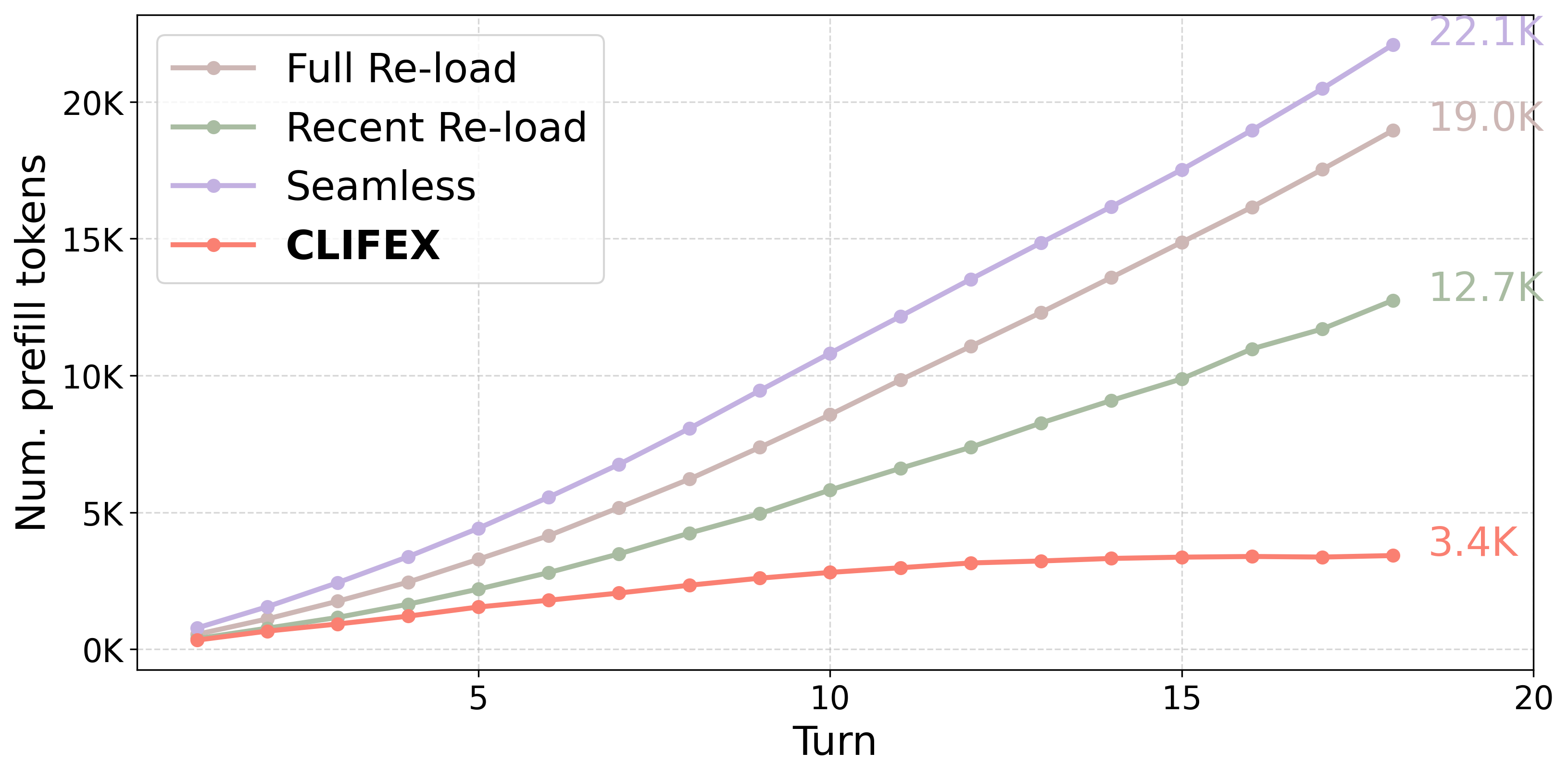}
    \caption{Main-task execution}
    \end{subfigure}
    \caption{
    \textbf{Turn-wise prefilled tokens for sub-task classification, and main-task and sub-task execution until each turns} in LLaMA3.1-Instruct (8B) on QReCC-Task+ dataset.} 
    \label{fig:exp_token_qrecc}
    \vspace{-0.2cm}
\end{figure}

\vspace{-0.3cm}
As shown in~\ref{fig:exp_token_qrecc}, we can see the similar trend of the prefill-token efficiency with TopioCQA-Task+ in sub-task classification, and sub and main task execution.

\section{Prompt for Evaluation Metric}
\vspace{-0.3cm}
\begin{tcolorbox}[fonttitle=\small\bfseries,
  fontupper=\scriptsize\ttfamily,
  colback=gray!5!white,
  colframe=gray!75!black,
  enhanced,
  sharp corners,
  boxrule=0.5pt,
  left=2pt, right=2pt, top=2pt, bottom=2pt,
  title=Correctness metric,
  breakable
]
\lstset{
  basicstyle=\scriptsize\ttfamily,
  breaklines=true,
  breakindent=0pt,      
  prebreak={},          
  postbreak={},         
  showstringspaces=false,
  columns=fullflexible, 
  keepspaces=true,
  xleftmargin=0pt,
  framexleftmargin=0pt
}
\begin{lstlisting}
Given the description for a dialogue, evaluates if the response can be considered as a correct response of the dialogue\n(labels: (0 : wrong, 0.5 : partial, 1 : correct)\n You should return only digit 0, 0.5, or 1. \nresponse: {response}\n detailed dialogue description: {answer}
\end{lstlisting}
\end{tcolorbox}

\begin{tcolorbox}[fonttitle=\small\bfseries,
  fontupper=\scriptsize\ttfamily,
  colback=gray!5!white,
  colframe=gray!75!black,
  enhanced,
  sharp corners,
  boxrule=0.5pt,
  left=2pt, right=2pt, top=2pt, bottom=2pt,
  title=Coherence metric,
  breakable
]
\lstset{
  basicstyle=\scriptsize\ttfamily,
  breaklines=true,
  breakindent=0pt,      
  prebreak={},          
  postbreak={},         
  showstringspaces=false,
  columns=fullflexible, 
  keepspaces=true,
  xleftmargin=0pt,
  framexleftmargin=0pt
}
\begin{lstlisting}
Each dialogue is multi-turn conversation between user and AI assistant. Assesses whether the response is naturally and coherently structured, connecting the prompt\n  (labels: 0 : not coherent, 0.5 : partially coherent, 1 : coherent).\n You should return only digit 0, 0.5, or 1.\nprompt: {prompt}\nresponse: {response}
\end{lstlisting}
\end{tcolorbox}

\section{Inference Prompt Templates}
\vspace{-0.3cm}
We also provide the prompt templates used for inference (sub-task classification, sub-task execution, main-task execution) in Fig.~\ref{fig:inf_main},~\ref{fig:inf_cls},~\ref{fig:inf_sub}. In the side-task, we designed that a large parts is shared across sub-tasks as shown in~\ref{fig:inf_sub}.

\section{Dataset}
\label{app:dataset}
\vspace{-0.3cm}
\subsection{Data Curation Prompt}
We fully provide the dataset curation prompt in Fig. ~\ref{fig:db_api},~\ref{fig:db_casual_1},~\ref{fig:db_casual_2},~\ref{fig:db_casual_3}.

\subsection{More Details}
In addition, for RAG and math problem solving turns, we use the original datasets themselves where math problems are randomly sampled for each conversational search (RAG) dialogue. The detailed process of data generation is as follows:
\begin{itemize}
    \item Select a multi-turn conversational search from TopiOCQA or QReCC as the conversation history. API call conversation is generated using the prompt of Fig.~\ref{fig:db_api}. The resulting API call turns are added to the conversation history. 
    \item On top of the conversational search and API call turns, the start of casual chat (Fig.~\ref{fig:db_casual_1} is generated. Then, the on-going turns are iteratively generated with Fig.~\ref{fig:db_casual_2}. Lastly, the finishing turn is generated via Fig.~\ref{fig:db_casual_3}. 
    \item Math solving problems are randomly selected from the GSM8K dataset, and added to the latest conversation history.

\end{itemize}
Finally, the curated multi-turn conversation is a series of RAG questions, multi-modal questions, casual chat (start, on-going, ending), and math solving. Notice that, as rewriting and retrieval augmented generation (RAG) is more challenging than other tasks, this kind of turns are about 40\% in each series of turns.

For fairness, we employed the LLaMA3.1-Instruct (70B) for the generated turns since we use the GPT4 for the evaluation. 

We provide examples of the generated turns in our datasets:
\begin{tcolorbox}[mypromptbox, title=API Call]
Q) Can you classify the organisms in this underwater image? [Image input]

A) Sure, I used the $\textrm{resnext101\_32x8d\_wsl}$ model. The organisms in the image include benthic crustaceans.
\end{tcolorbox}

\begin{tcolorbox}[mypromptbox, title=Casual Chat]
Q) It’s finally sunny outside after days of rain!

A) That’s great news, isn’t it? A bit of sunshine always lifts up our mood

Q) I’m so glad I can finally take a walk 

A) Me too, there’s nothing like feeling the warm sum on your skin to boost your energy levels!

…

Q) Thanks for chatting with me about the lovely weather, have a wonderful rest of your day!

A) It was great chatting with you too, enjoy your coffee and time at the park!
\end{tcolorbox}

\clearpage

\begin{figure*}[h]
\section{Prompt Templates}
\label{app:template}

\centering
\begin{tcolorbox}[fonttitle=\small\bfseries,
  fontupper=\scriptsize\ttfamily,
  colback=gray!5!white,
  colframe=gray!75!black,
  enhanced,
  sharp corners,
  boxrule=0.5pt,
  left=2pt, right=2pt, top=2pt, bottom=2pt,
  title=Prompt template for main task,
  width=\textwidth,
  breakable
]
\lstset{
  basicstyle=\scriptsize\ttfamily,
  breaklines=true,
  breakindent=0pt,
  showstringspaces=false,
  columns=fullflexible,
  keepspaces=true,
  xleftmargin=0pt,
  framexleftmargin=0pt
}
\begin{lstlisting}
A multi-trun or single-turn conversation between user and AI assistant is given. \
Answer the last user question. Ensure that only answer the last user question. If a context is provided for the last question, use the context as the major source for the answer. \
Answer should be less than 50 words. Do not output conversation history or other unnecessary words as "Here is..." 

#question1: {question_1}
#answer1: {answer_1}
.
.
.
#question{n}: {question_n}
\end{lstlisting}
\end{tcolorbox}
\caption{Prompt template for main-task execution.}
\label{fig:inf_main}
\end{figure*}
\vspace{-2.0cm}


\begin{figure*}[h]
\centering
\begin{tcolorbox}[fonttitle=\small\bfseries,
  fontupper=\scriptsize\ttfamily,
  colback=gray!5!white,
  colframe=gray!75!black,
  enhanced,
  sharp corners,
  boxrule=0.5pt,
  left=2pt, right=2pt, top=2pt, bottom=2pt,
  title=Prompt template for sub-task clssi. (per-task binary),
  width=\textwidth,
  breakable
]
\lstset{
  basicstyle=\scriptsize\ttfamily,
  breaklines=true,
  breakindent=0pt,
  showstringspaces=false,
  columns=fullflexible,
  keepspaces=true,
  xleftmargin=0pt,
  framexleftmargin=0pt
}
\begin{lstlisting}
Your new goal: for the target last user question, discern if the following sub-task is required or not.

Answer format: Ensure to provide your answer as ONLY "Yes" or "No" without adding unnecessary words or reasons, descriptions. 
i.e. "#Answer: ["Yes" or "No"]" 

Sub-task: 
{Sub-task description}.
\end{lstlisting}
\end{tcolorbox}
\caption{Prompt template for turns for answering in sub-task.}
\label{fig:inf_cls}
\end{figure*}
\vspace{-2.0cm}

\begin{figure*}[h]
\centering
\begin{tcolorbox}[fonttitle=\small\bfseries,
  fontupper=\scriptsize\ttfamily,
  colback=gray!5!white,
  colframe=gray!75!black,
  enhanced,
  sharp corners,
  boxrule=0.5pt,
  left=2pt, right=2pt, top=2pt, bottom=2pt,
  title=Prompt template for sub-task execution,
  width=\textwidth,
  breakable
]
\lstset{
  basicstyle=\scriptsize\ttfamily,
  breaklines=true,
  breakindent=0pt,
  showstringspaces=false,
  columns=fullflexible,
  keepspaces=true,
  xleftmargin=0pt,
  framexleftmargin=0pt
}
\begin{lstlisting}
Note that, now you are a helpful, respectful and honest assistant for a following sub-task.

Sub-task: 
{Sub-task description}.
\end{lstlisting}
\end{tcolorbox}
\caption{Prompt template for turns for sub-task execution (ours).}
\label{fig:inf_sub}
\end{figure*}
\vspace{-2.0cm}

\begin{figure*}[t]
\centering
\begin{tcolorbox}[fonttitle=\small\bfseries,
  fontupper=\scriptsize\ttfamily,
  colback=gray!5!white,
  colframe=gray!75!black,
  enhanced,
  sharp corners,
  boxrule=0.5pt,
  left=2pt, right=2pt, top=2pt, bottom=2pt,
  title=Dataset: Prompt template for API calling turn,
  width=\textwidth,
  breakable
]
\lstset{
  basicstyle=\scriptsize\ttfamily,
  breaklines=true,
  breakindent=0pt,
  showstringspaces=false,
  columns=fullflexible,
  keepspaces=true,
  xleftmargin=0pt,
  framexleftmargin=0pt
}
\begin{lstlisting}
You are a helpful conversation simulator between User and AI assistant. \
Given a series of previous conversation (question and answers are seperated by [SEP]), make a follow-up user interaction and AI
assistant response based on a hypothetical image.

Assume the image has been processed by a vision-language API (e.g., LLaVA), and the API has returned a detailed textual image description.

Your task is to generate:
1. A realistic **LLaVA-style textual image description sentences**, which clearly and explicitly mentions visual content.
2. A **casual user question** that refers to the image
3. An **AI assistant answer** that responds to the question based only on the given textual description.

---

### Guidelines

- The **user question** must:
  - Explicitly or implicitly refer to the image.
  - Be casual and natural in tone.

- The **assistant answer** must:
  - Use only information that is **present** in the image description.
  - Avoid interpretation, speculation, or reasoning.

---

### Format:

##Textual Image Description: [Generated description]  
##User Question: [User question that can be answered with no inference, just by reading the description] 
##AI Assistant Answer: [Answer using only what is stated in the description]

---

### Example:

##Textual Image Description: A small white dog is sitting on a blue couch with a red blanket draped over one side. A green ball lies on the floor in front of the couch, and sunlight is coming in through a nearby window.  
##User Question: What color is the couch in this image?  
##AI Assistant Answer: The couch is blue.

---

Now generate more examples in this format, strictly following the instructions above.

### Previous Conversation: {conversation_history}

Output:
\end{lstlisting}
\end{tcolorbox}
\caption{Prompt template: data curation for API call turns.}
\label{fig:db_api}
\end{figure*}

\begin{figure*}[t]
\centering
\begin{tcolorbox}[fonttitle=\small\bfseries,
  fontupper=\scriptsize\ttfamily,
  colback=gray!5!white,
  colframe=gray!75!black,
  enhanced,
  sharp corners,
  boxrule=0.5pt,
  left=2pt, right=2pt, top=2pt, bottom=2pt,
  title= Dataset: Prompt template for start of casual conversation,
  width=\textwidth,
  breakable
]
\lstset{
  basicstyle=\scriptsize\ttfamily,
  breaklines=true,
  breakindent=0pt,
  showstringspaces=false,
  columns=fullflexible,
  keepspaces=true,
  xleftmargin=0pt,
  framexleftmargin=0pt
}
\begin{lstlisting}
You are a helpful conversation simulator between User and AI assistant. \
Given a series of previous conversation (question and answers are separated by [SEP]), make a follow-up user interaction and AI assistant response. \
In the previous conversation, to answer the user questions, information retrieval or API calls are required. \
However, from now on, it's different. The follow-up user interaction should be non-knowledge intensive and not necessarily in the form of a question. \
It should be a casual chat, focusing on topics such as user's daily life or sentiment. \
The user interaction can be a statement, a greeting, or any other form of casual conversation starter. 

The user interaction and AI assistant response should consist of concise one or two sentences.

Make a single turn of user interaction and AI assistant response where each of them is seperated by [SEP]:
[User interaction] [SEP] [AI assistant response]

### Previous Conversation:
{conversation_history}

Output:
\end{lstlisting}
\end{tcolorbox}
\caption{Prompt template: data curation for starting casual chat turns.}
\label{fig:db_casual_1}
\end{figure*}

\begin{figure*}[t]
\centering
\begin{tcolorbox}[fonttitle=\small\bfseries,
  fontupper=\scriptsize\ttfamily,
  colback=gray!5!white,
  colframe=gray!75!black,
  enhanced,
  sharp corners,
  boxrule=0.5pt,
  left=2pt, right=2pt, top=2pt, bottom=2pt,
  title=Dataset: Prompt template for on-going casual conversation,
  width=\textwidth,
  breakable
]
\lstset{
  basicstyle=\scriptsize\ttfamily,
  breaklines=true,
  breakindent=0pt,
  showstringspaces=false,
  columns=fullflexible,
  keepspaces=true,
  xleftmargin=0pt,
  framexleftmargin=0pt
}
\begin{lstlisting}
You are a helpful conversation simulator between User and AI assistant. \
Given a series of previous conversation (question and answers are separated by [SEP]), make a follow-up user interaction and AI assistant response. \
Similar to the last conversation, the follow-up user interaction and AI assistant response should be casual chat with no need of knowledge retrieval. \
Namely, it should be a casual chat, focusing on topics such as user's daily life or sentiment.   \
Also, it is not necessarily in the form of a question. \

The user interaction and AI assistant response should consist of concise one or two sentences.

Make a single turn of user interaction and AI assistant response where each of them is seperated by [SEP]:
[User interaction] [SEP] [AI assistant response]

### Previous Conversation:
{conversation_history}

Output:
\end{lstlisting}
\end{tcolorbox}
\caption{Prompt template: data curation for on-going casual chat turns.}
\label{fig:db_casual_2}
\end{figure*}

\begin{figure*}[t]
\centering
\begin{tcolorbox}[fonttitle=\small\bfseries,
  fontupper=\scriptsize\ttfamily,
  colback=gray!5!white,
  colframe=gray!75!black,
  enhanced,
  sharp corners,
  boxrule=0.5pt,
  left=2pt, right=2pt, top=2pt, bottom=2pt,
  title= Dataset-Prompt template for finishing casual conversation,
  width=\textwidth,
  breakable
]
\lstset{
  basicstyle=\scriptsize\ttfamily,
  breaklines=true,
  breakindent=0pt,
  showstringspaces=false,
  columns=fullflexible,
  keepspaces=true,
  xleftmargin=0pt,
  framexleftmargin=0pt
}
\begin{lstlisting}
You are a helpful conversation simulator between User and AI assistant. \
Given a series of previous conversation (question and answers are separated by [SEP]), make the final turn of user interaction and AI assistant response. \
Crucially, the final turn must imply that the conversation is ending.
The user interaction and AI assistant response should consist of concise one or two sentences.

Make a single turn of questions and answers where each of them is seperated by [SEP]:
[User interaction] [SEP] [AI assistant response]

### Previous Conversation:
{conversation_history}

Output:
\end{lstlisting}
\end{tcolorbox}
\caption{Prompt template: data curation for finishing casual chat turns.}
\label{fig:db_casual_3}
\end{figure*}